\documentclass{article}

\usepackage[main, final]{neurips_2026}
\usepackage{graphicx}
\usepackage{amsmath}
\usepackage{booktabs}   %
\usepackage{multirow}   %
\usepackage{adjustbox}  %
\usepackage{siunitx}    %
\usepackage{todonotes}

\usepackage[utf8]{inputenc} %
\usepackage[T1]{fontenc}    %
\usepackage{hyperref}       %
\usepackage{url}            %
\usepackage{booktabs}       %
\usepackage{amsfonts}       %
\usepackage{nicefrac}       %
\usepackage{microtype}      %
\usepackage{xcolor}         %
\usepackage{array}
\usepackage{flafter}   %
\usepackage{placeins}  %
\usepackage{pifont}
\usepackage{wrapfig}

\title{3DROID: A Renderable 3D Gaussian Dataset with Measured Per-Scene Reliability}

\author{%
  Wonguen Cho$^{1}$ \qquad Junhoo Lee$^{2}$ \qquad Nojun Kwak$^{1}$ \\[3pt]
  \normalfont $^{1}$Seoul National University \qquad $^{2}$KAIST \\
  \normalfont \texttt{\{wonguen.cho, nojunk\}@snu.ac.kr} \qquad \texttt{junhoo.lee@kaist.ac.kr}
}

\begin{document}

\maketitle

\begin{abstract}
Robot manipulation models primarily reason from 2D observations while acting in the 3D physical world. 
To bridge this gap, recent work has augmented robot data with geometric priors such as depth, point clouds, and 3D trajectories, while renderable 3D Gaussian representations provide another promising form of 3D supervision. 
However, 3DGS representation is designed mainly for photometric fidelity and may not preserve real-world metric scale, particularly when the supplied camera extrinsics are unreliable. 
We study the effect of extrinsic reliability and pose conditioning on feed-forward 3DGS, and propose a calibration-aware pipeline that anchors reconstructed scenes to the robot's metric workspace.
Our experiments show that pose conditioning improves novel-view fidelity, while its geometric benefit depends on the reliability of the injected extrinsics.
Using this pipeline, we present a renderable, metric-pose-anchored dataset with scene-level reliability information for robot manipulation research. Our dataset is available at
\\\url{https://huggingface.co/datasets/wonguen/3DROID}
\end{abstract}

\section{Introduction}

Robot manipulation is inherently geometric: grasping and placement require
reasoning about objects, free space, and motion in 3D. Explicit 3D
representations improve spatial reasoning over 2D observations~\cite{lift3dvla2024, yang2025fp33dfoundationpolicy, yang2026robo3r,zhang2025vla}, motivating recent 3D annotations and renderable robot scenes
~\cite{dong2025embodiedmae,wang2026gensplat,zhang2026robosnaponeshotrealtosimscene}.
Unlike point- or token-based annotations, renderable scenes provide a view-consistent appearance model from which observations and occlusions can be generated at new camera poses, enabling visual augmentation and simulation without additional data collection~\cite{zhang2026robosnaponeshotrealtosimscene}.
Such novel-view augmentation has also been shown to improve policy robustness
to unseen camera poses~\cite{wang2026gensplat}.

Constructing these representations at dataset scale favors feed-forward reconstruction. 
Classical 3DGS optimizes each scene independently from its
available observations, making sparse robot views underconstrained and requiring repeated optimization over thousands of episodes
~\cite{kerbl3Dgaussians}. 
Feed-forward variants instead amortize reconstruction
across training scenes: their learned priors support inference from limited views, while a single forward pass avoids repeated per-scene optimization ~\cite{jiang2025anysplat,veicht2026zipsplat,yonosplat,ye2024noposplat}.
This makes them a practical basis for constructing large-scale renderable robot datasets.

These advantages address sparsity and computational cost, but not physical reliability. 
Both classical and feed-forward 3DGS are developed primarily for
photorealistic novel-view synthesis. 
For manipulation, however, synthesized observations are useful only if they remain consistent with the metric workspace in which robot states and actions are defined~\cite{yang2026robo3r}.
A reconstruction may preserve appearance while silently moving surfaces or changing scene scale.
Moreover, camera calibrations in real-world robot datasets are not uniformly reliable, 
so shipped poses cannot simply be treated as ground truth. 
Whether the resulting scenes preserve metric scale therefore remains unclear, 
and calibration must be validated independently of the reconstructor.

\begin{table*}[!t]
  \centering
  \small
  \newcommand{\yes}{\ding{51}}
  \newcommand{\no}{\ding{55}}
  \begin{tabular}{l cc}
    \toprule
    Per-scene asset & DROID~\cite{khazatsky2024droid} & \textbf{3DROID} \\
    \midrule
    Stereo RGB video                 & \yes & \no \\
    Robot joint state                & \yes & \no \\
    Camera intrinsics$^{\dag}$                & \yes & \yes \\
    Shipped extrinsics$^{\dag}$               & \yes & \yes \\
    Refined extrinsics$^{\ddag}$   & \no  & \yes \\
    \midrule
    Renderable 3D Gaussians          & \no  & \yes \\
    Metric depth maps                & \no  & \yes \\
    Coverage masks                   & \no  & \yes \\
    Calibration-gate verdict         & \no  & \yes \\
    Per-scene reliability envelope   & \no  & \yes \\
    \bottomrule
  \end{tabular}
  \caption{
Data availability in DROID and 3DROID. 3DROID adds renderable reconstructions
and measured reliability to 114 DROID scenes; source videos and robot states
remain in DROID and are joined by episode name.
$^{\dagger}$Intrinsics and shipped extrinsics are from DROID.
$^{\ddagger}$PointWorld provides the refined left-eye poses; the corresponding
right-eye poses are derived from the known within-rig stereo transform. All
camera parameters are packaged in a common convention, with no additional
calibration performed by us.
}
  \label{tab:dataset_delta}
\end{table*}

Prior work addresses parts of this problem from complementary directions.
Geometry-aware Gaussian methods introduce surface constraints, stereo cues, or 3D priors to improve geometric stability~\cite{Huang2DGS2024, wolf2024gsmesh,wang2026gensplat}, while camera-refinement methods align visual observations with robot geometry~\cite{huang2026pointworld}. 
Yet these model- and calibration-level solutions do not by themselves specify how reliability should be communicated when reconstructions are released as data. 
This leaves a dataset-level question: how can calibration suitability and reconstruction reliability be measured and exposed for each scene?

In this paper, we therefore build \textbf{3DROID} by reconstructing DROID~\cite{khazatsky2024droid} with pose-conditioned feed-forward 3DGS and
separating photometric fidelity from physical consistency. 
Building on robot--depth alignment used for camera refinement~\cite{huang2026pointworld}, we validate calibration before reconstruction and separately assess geometric reliability afterward, together with measurement coverage.
This separates calibration suitability from reconstruction reliability.
We therefore release measured support rather than claim geometric ground truth or downstream policy gains.
Table~\ref{tab:dataset_delta} summarizes the scene-level assets added by 3DROID and the source modalities that remain in DROID.

Our contributions are threefold:
\begin{itemize}
    \item We release \textbf{3DROID}, a DROID-derived dataset of renderable 3D Gaussian scenes with per-scene calibration and reconstruction-reliability metadata.
    \item We empirically show that metric pose conditioning anchors reconstruction scale to the physical workspace and yields median per-scene PSNR gains of $+0.73$ to $+2.73$\,dB at a fixed extrinsic source.
    \item We repurpose robot--depth consistency as a coverage-aware reconstruction evaluation protocol that compares rendered 3DGS depth with classical stereo and kinematically localized robot geometry.
\end{itemize}

\section{Related works}
 
 \begin{figure}
     \centering
\includegraphics[width=1.0\linewidth]{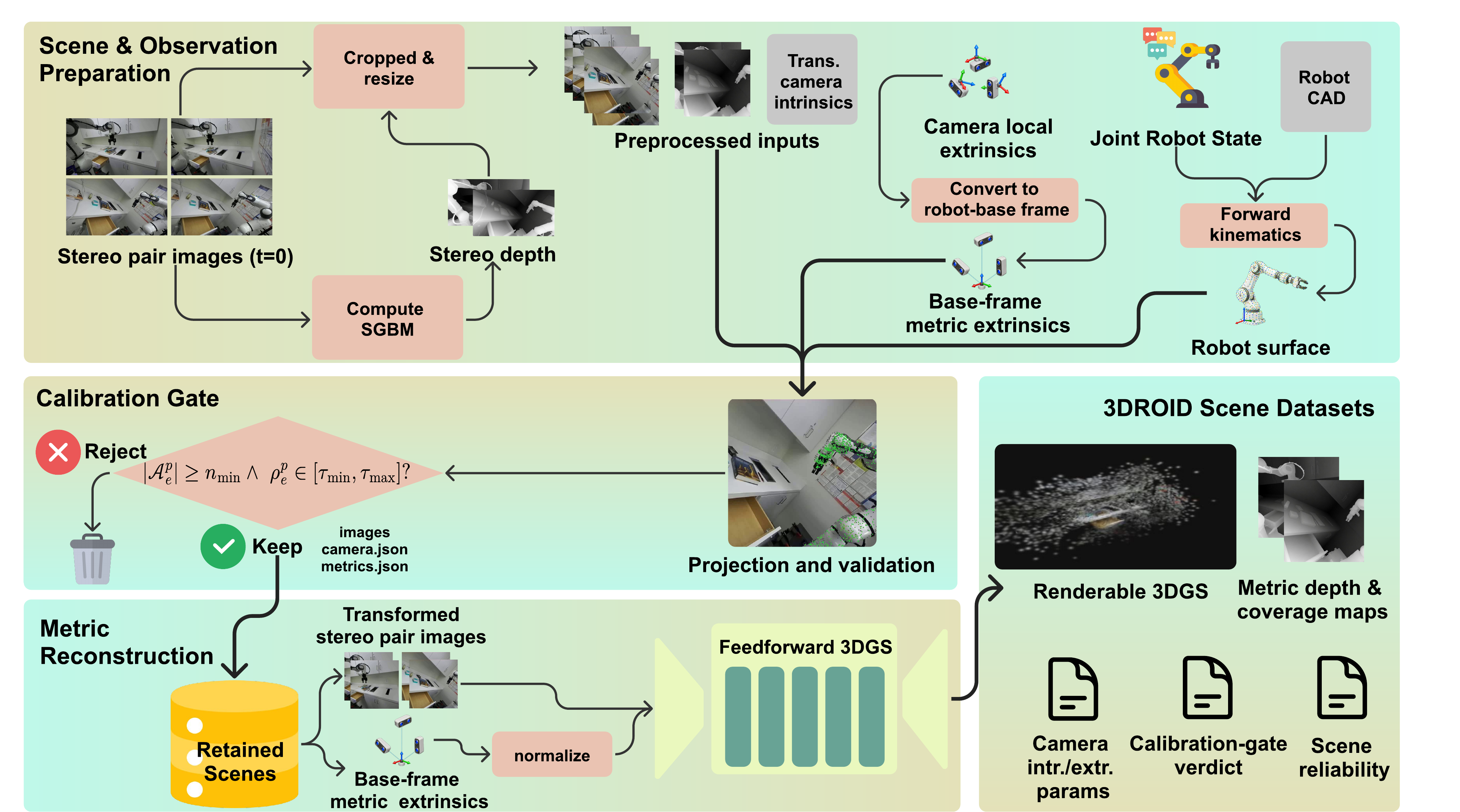} 
    \caption{\textbf{Overview of the 3DROID pose-conditioned construction pipeline.}
The figure summarizes the scene preparation, calibration gate, reconstruction,
and dataset outputs detailed in Section~\ref{sec:robo}.}
    \label{fig:pipeline}     
  \end{figure}
  
\paragraph{3D representations for robotics.}
Explicit 3D observations improve spatial reasoning for manipulation. DP3 demonstrated this with sparse point cloud, and FP3 scaled point-cloud pretraining to DROID~\cite{yang2025fp33dfoundationpolicy,Ze2024DP3}.
DROID-3D added depth and point clouds, 
while 4D-VLA and Lift3D-VLA incorporated temporal geometry~\cite{dong2025embodiedmae,lift3dvla2024,zhang2025vla}.
PointWorld and Robo3R further improved calibration and robot-frame alignment~\cite{huang2026pointworld, yang2026robo3r}. Yet these point- or token-based outputs lack a view-consistent appearance model. Unlike point observations, renderable scenes can synthesize view-consistent observations from new camera poses, expose viewpoint-dependent visibility, and support data augmentation or visual simulation without recollecting demonstration. 
RoboSnap builds layered mesh--Gaussian scenes from a single image, but does
not provide measured per-scene reconstruction reliability~\cite{zhang2026robosnaponeshotrealtosimscene}. 
This leaves open how to construct renderable robot scenes while explicitly measuring their metric alignment and reconstruction reliability.

\paragraph{Feedforward Gaussian Splatting.}
Whereas per-scene 3DGS requires time-consuming optimization~\cite{kerbl3Dgaussians}, 
feed-forward methods leverage learned priors to reconstruct scenes from sparse views in a single pass~\cite{charatan23pixelsplat,jiang2025anysplat}. 
Regardless of reconstruction strategy, 3DGS is designed primarily for photorealistic novel-view rendering.
Yet photorealism and manipulation readiness are different objectives: novel-view fidelity does not guarantee metric scale, robot-frame alignment, or reliable geometry.
Pose-conditioned PixelSplat, MVSplat, and DepthSplat improve matching and depth estimation but assume reliable calibration~\cite{charatan23pixelsplat,chen2024mvsplat,xu2024depthsplat}. NoPoSplat and AnySplat remove pose inputs, while YoNoSplat handles variable posed or unposed inputs and ZipSplat compresses the representation~\cite{jiang2025anysplat, veicht2026zipsplat, yonosplat, ye2024noposplat}. GenSplat applies this pose-free paradigm to robotic view augmentation, but evaluates DROID reconstruction only qualitatively without testing robot-frame alignment~\cite{wang2026gensplat}. We instead audit calibration and evaluate rendering separately from stereo-supported geometry.

\section{Robot-Anchored 3D Reconstruction}
\label{sec:robo}
\subsection{Pose-Free Feed-Forward Gaussian Reconstruction}

Feed-forward Gaussian reconstructors predict a 3D Gaussian scene from a set
of source-view inputs. For scene $e$, let $\mathcal X_e$ denote the native
pose-free input of a backbone. The reconstructor predicts
\begin{equation}
    \widehat{\mathcal G}_e = F_\theta(\mathcal X_e),
    \qquad
    \widehat I_{et}
    =
    \mathcal R\!\left(
        \widehat{\mathcal G}_e,\xi_{et}
    \right),
    \label{eq:pose_free_reconstruction}
\end{equation}
where $\xi_{et}$ denotes a target camera configuration and
$\widehat I_{et}$ is the corresponding rendered image. These models are
trained to minimize target-view image discrepancy,
\begin{equation}
    \theta^\star
    =
    \arg\min_\theta
    \mathbb E_{e,t}
    \left[
        d_{\mathrm{img}}
        \left(
            \widehat I_{et}, I_{et}
        \right)
    \right],
    \label{eq:novel_view_objective}
\end{equation}
where $d_{\mathrm{img}}$ denotes the model-specific image-space training
objective. Thus, pose-free feed-forward reconstruction learns a scene whose
renderings agree with images from held-out viewpoints.

\subsection{3D Reconstruction from a Robot Manipulation Perspective}

Image-space supervision does not by itself provide a global metric reference.
Let
\begin{equation}
    \mathbf S(\mathbf x)
    =
    s\mathbf R\mathbf x+\mathbf t,
    \qquad
    \mathbf S\in\mathrm{Sim}(3),
    \quad
    s>0,
    \quad
    \mathbf R\in\mathrm{SO}(3).
    \label{eq:sim3_transform}
\end{equation}
Applying the same global similarity transform to a scene and its camera
configuration preserves the rendered images:
\begin{equation}
    \mathcal R(\widehat{\mathcal G}_e,\mathcal C_e)
    =
    \mathcal R\!\left(
        \mathbf S\!\cdot\!\widehat{\mathcal G}_e,
        \mathbf S\!\cdot\!\mathcal C_e
    \right).
    \label{eq:sim3_ambiguity}
\end{equation}
Consequently, image-only reconstruction is determined only up to a global
$\mathrm{Sim}(3)$ transformation.

Robot manipulation instead requires a metric reconstruction anchored to the
robot base frame $\mathsf B$. Let
$\mathbf P_{ei}^{\mathsf B}$ be a robot-surface point obtained by applying
forward kinematics to joint state $\mathbf q_e$ and CAD point $\mathbf c_i$,
\begin{equation}
    \mathbf P_{ei}^{\mathsf B}
    =
    \operatorname{FK}(\mathbf q_e;\mathbf c_i).
    \label{eq:robot_surface_points}
\end{equation}
For the ideal base-to-camera extrinsic
$(\mathbf R^\star_{ev},\mathbf t^\star_{ev})$, this physical point projects
to
\begin{equation}
\begin{aligned}
    \mathbf x^\star_{eiv}
    &=
    \mathbf R^\star_{ev}\mathbf P_{ei}^{\mathsf B}
    +
    \mathbf t^\star_{ev}, \\
    \mathbf u^\star_{eiv}
    &=
    \pi\!\left(
        \mathbf K_{ev}\mathbf x^\star_{eiv}
    \right),
    \qquad
    z^\star_{eiv}
    =
    [\mathbf x^\star_{eiv}]_z, \\
    D^{\mathrm{true}}_{ev}
    \left(
        \mathbf u^\star_{eiv}
    \right)
    &=
    z^\star_{eiv}.
\end{aligned}
\label{eq:ideal_metric_consistency}
\end{equation}
This condition defines the desired metric anchor: robot geometry expressed in
$\mathsf B$ must agree with the corresponding camera-depth observation.

\subsection{Dataset Curation via Calibration Consistency Validation}

However, DROID camera poses can be noisy~\cite{huang2026pointworld}, so even a refined pose cannot be assumed to be a reliable metric reference. We therefore replace the unavailable ideal quantities in Eq.~\ref{eq:ideal_metric_consistency} with classical stereo depth and validate each candidate pose source $p$ at projected robot points. For each valid
robot-point projection, we compute
\begin{equation}
\begin{aligned}
    \mathbf x^p_{eiv}
    &=
    \mathbf R^p_{ev}\mathbf P_{ei}^{\mathsf B}
    +
    \mathbf t^p_{ev}, \\
    \mathbf u^p_{eiv}
    &=
    \pi\!\left(
        \mathbf K_{ev}\mathbf x^p_{eiv}
    \right),
    \qquad
    z^p_{eiv}
    =
    [\mathbf x^p_{eiv}]_z, \\
    r^p_{eiv}
    &=
    \frac{
        D^{\mathrm{st}}_{ev}
        \left(
            \mathbf u^p_{eiv}
        \right)
    }{
        z^p_{eiv}
    }.
\end{aligned}
\label{eq:observable_metric_proxy}
\end{equation}
Here $D^{\mathrm{st}}_{ev}$ is the stereo depth map and
$\mathcal A_e^p$ contains projections that are in crop and have valid stereo
depth. We aggregate the partial observations using the robust scene-level
statistic
\begin{equation}
    \rho_e^p
    =
    \operatorname{median}_{(i,v)\in\mathcal A_e^p}
    r^p_{eiv},
    \label{eq:scene_level_metric_proxy}
\end{equation}
and retain the scenes satisfying
\begin{equation}
    \mathcal E_{\mathrm{valid}}^p
    =
    \left\{
        e\;\middle|\;
        |\mathcal A_e^p|\ge n_{\min},
        \quad
        \tau_{\min}\le\rho_e^p\le\tau_{\max}
    \right\}.
    \label{eq:valid_scene_set}
\end{equation}
For a pose-compatible reconstructor, the validated camera configuration is
then used for pose-conditioned reconstruction,
\begin{equation}
    \widehat{\mathcal G}^{\mathsf B}_e
    =
    F_\theta\!\left(
        \mathcal X_e,\mathcal C_e^p
    \right),
    \qquad
    e\in\mathcal E_{\mathrm{valid}}^p,
    \label{eq:pose_conditioned_reconstruction}
\end{equation}
where
$\mathcal C_e^p=
\{ \mathbf K_{ev},
[\mathbf R^p_{ev}\mid\mathbf t^p_{ev}] \}_{v\in\mathcal V_e}$.

\begin{table*}[t]
  \centering
  \small
  \setlength{\tabcolsep}{3.5pt}
  \renewcommand{\arraystretch}{0.95}
  \begin{tabular}{ll rrr}
    \toprule
    Backbone & Method
    & PSNR$\uparrow$
    & SSIM$\uparrow$
    & LPIPS$\downarrow$ \\
    \midrule

    \multirow{2}{*}{YoNoSplat~\cite{yonosplat}}
    & Baseline
    & 12.57 & 0.3582 & 0.5588 \\
    & Ours
    & \textbf{15.81} & \textbf{0.5105} & \textbf{0.2592} \\

    \midrule
    \multirow{2}{*}{ZipSplat~\cite{veicht2026zipsplat}}
    & Baseline
    & 12.75 & 0.4233 & 0.5467 \\
    & Ours
    & \textbf{16.27} & \textbf{0.5242} & \textbf{0.2979} \\

    \bottomrule
  \end{tabular}
  \caption{
  Photometric fidelity from three-view reconstructions on 110 scenes.
  Baseline is pose-free and evaluated with shipped DROID extrinsics. Ours uses
  refined extrinsics with pose conditioning. Bold indicates the better result
  within each backbone.
  }
  \label{tab:photometric}
\end{table*}

\section{3DROID}

\subsection{Configuration}

DROID is widely used in recent manipulation research~\cite{khazatsky2024droid}. 
We group its roughly 35k successful episodes by $(\mathrm{lab},\mathrm{serial}_1,\mathrm{serial}_2,\mathrm{date})$ and sample one episode per group, 
yielding 520 scene groups. From these scenes, 225 scenes contain the four stereo videos and trajectory record required by our gate. 
Applying the calibration gate in Eq.~\ref{eq:valid_scene_set} with $n_{\min}=30$ valid projections and
$(\tau_{\min},\tau_{\max})=(0.85,1.15)$ retains 114 scenes under the refined
pose source and 80 under the shipped one. We release the former.

As summarized in Table~\ref{tab:dataset_delta}, \textbf{3DROID} augments each
retained episode with a pose-conditioned ZipSplat reconstruction~\cite{veicht2026zipsplat}, rendered depth and coverage maps, camera
parameters, and per-scene reliability measurements. We perform no new camera
calibration: intrinsics and shipped extrinsics come from DROID, while
PointWorld provides the refined left-eye extrinsics~\cite{huang2026pointworld}, and corresponding right-eye poses are derived using
the original stereo transform. All parameters are packaged in a common
camera-to-world convention, and source videos and robot states remain
accessible through the DROID episode key.

Four released scenes contain all-zero DROID intrinsics and yield no valid
evaluation measurements, leaving 110 scenes for photometric evaluation. One
additional scene uses per-serial median intrinsics from 501 other DROID
episodes and is retained. All five cases are explicitly flagged. Geometric
comparisons involving shipped extrinsics use 109 scenes because one further
scene lacks valid robot support, whereas comparisons within refined
extrinsics use 110.

\subsection{Evaluation}

\paragraph{Baselines}
We evaluate ZipSplat~\cite{veicht2026zipsplat} and
YoNoSplat~\cite{yonosplat} at their native resolutions. Baseline is pose-free
with DROID extrinsics, whereas Ours is pose-conditioned with refined rig
extrinsics. We report geometric results for both backbones to assess whether
the same trend holds across models. In pose-free runs, the backbone reconstructs the scene from RGB alone and
predicts its own camera frame. The selected extrinsics are applied only
after reconstruction to register this predicted frame and place the held-out
camera for evaluation. They are not network inputs.

\paragraph{Settings}
Photometric evaluation reconstructs from \texttt{ext1L}, \texttt{ext1R}, and
\texttt{ext2L}, and reports PSNR, SSIM, and LPIPS on held-out \texttt{ext2R}.
Because the feed-forward backbones use the first input view as their  frame coordinate, we place \texttt{ext1L} first and hold out \texttt{ext2R}, which has the largest viewpoint separation from this anchor and thus provides a more discriminative evaluation~\cite{jiang2025anysplat,wang2025vggt}.
Geometric evaluation separately reconstructs from all four exterior views and reports robot-surface depth error and dense disagreement with classical stereo~\cite{hirschmuller2008stereo}, both with coverage. 
The wrist stream is excluded from both evaluations.
Dense stereo applies consistency and low-texture filtering and is treated as input-view agreement, not ground truth. 
All reported p-values are from two-sided Wilcoxon signed-rank tests on paired per-scene differences, with the scene as the unit of variability.

\subsubsection{Quantitative Results}
\begin{table*}[t]
  \centering
  \small
  \setlength{\tabcolsep}{4pt}
  \renewcommand{\arraystretch}{0.95}
  \begin{tabular}{ll rrr}
    \toprule
    \multirow{2}{*}{Backbone}
    & \multirow{2}{*}{Variant}
    & \multicolumn{2}{c}{Robot-referenced}
    & \multirow{2}{*}{\shortstack{Stereo Disagr.\\(cm)$\downarrow$}} \\
    \cmidrule(lr){3-4}
    & & Proj. Cov. & Err. (cm)$\downarrow$ & \\
    \midrule

    \multirow{4}{*}{YoNoSplat~\cite{yonosplat}}
    & Baseline
      & 0.155 & 20.38 & 10.70 \\
    & Baseline + pose inj.
      & 0.155 & \textbf{18.53} & \textbf{9.93} \\
    \cmidrule(lr){2-5}
    & Baseline + refined ext.
      & 0.167 & 7.08 & \textbf{7.52} \\
    & Ours
      & 0.167 & \textbf{6.92} & 7.54 \\

    \midrule
    \multirow{4}{*}{ZipSplat~\cite{veicht2026zipsplat}}
    & Baseline
      & 0.155 & \textbf{19.71} & 16.40 \\
    & Baseline + pose inj.
      & 0.155 & 30.58 & \textbf{11.80} \\
    \cmidrule(lr){2-5}
    & Baseline + refined ext.
      & 0.167 & 8.85 & 8.99 \\
    & Ours
      & 0.167 & \textbf{8.07} & \textbf{8.04} \\

    \bottomrule
  \end{tabular}

\caption{
\textbf{Geometric consistency of four-view reconstructions.} Within each backbone,
the upper and lower pairs use shipped ($n=109$) and refined ($n=110$)
extrinsics, respectively. The extra scene is unmeasurable with shipped
extrinsics because all robot queries project off-frame. The second row of each
pair injects poses. Because the extrinsic sources define different evaluation
geometries, comparisons are made only within each pair.
}

  \label{tab:geometric}
\end{table*}

\begin{figure*}[t]

 \centering
 \includegraphics[width=0.8\linewidth]{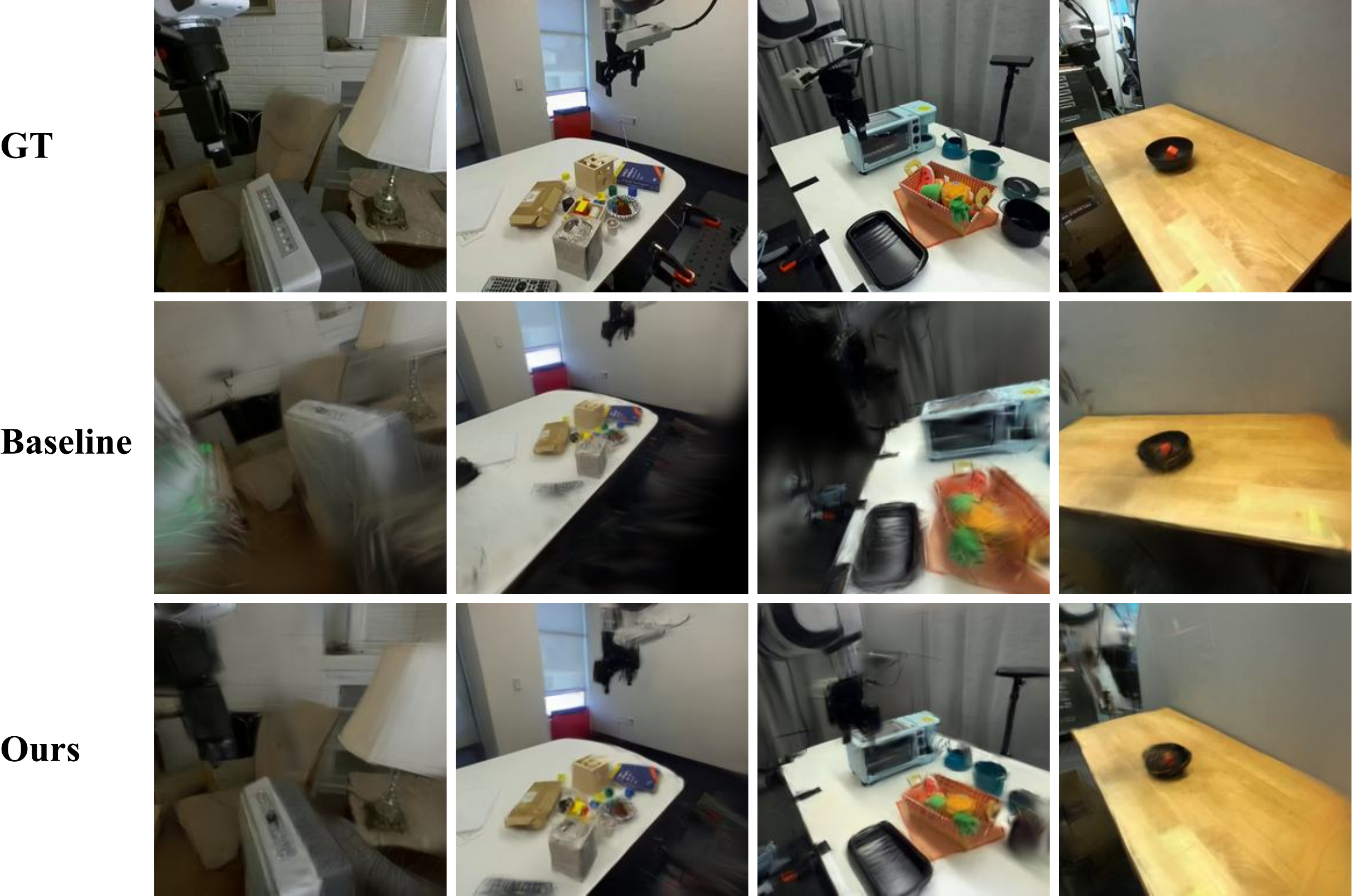}
  \caption{
  \textbf{Qualitative held-out rendering with ZipSplat.}
  Each column shows a different scene. Rows show the ground-truth
  \texttt{ext2R} image, the pose-free baseline using shipped DROID
  extrinsics, and our pose-conditioned reconstruction using
  PointWorld-refined extrinsics. All reconstructions use the remaining three
  exterior views.
  }
  \label{fig:qual}
\end{figure*}

\paragraph{Photometric Fidelity.}
Table~\ref{tab:photometric} compares our pose-conditioned, refined-extrinsic
configuration with the default pose-free baseline on held-out
\texttt{ext2R}. Across 110 scenes, PSNR increases from $12.57$ to $15.81$
($+3.24$\,dB) for YoNoSplat and from $12.75$ to $16.27$ ($+3.52$\,dB) for
ZipSplat. SSIM also increases and LPIPS decreases for both backbones. This
consistent trend shows that the photometric gain is not specific to the
ZipSplat release backbone. Because this end-to-end comparison changes both
the extrinsic source and pose conditioning, Table~\ref{tab:photometric_ablation}
separates their individual effects.

\paragraph{Geometric Fidelity.}
Table~\ref{tab:geometric} reports two complementary measures of geometric consistency.
\emph{Robot Projection Coverage} reports the fraction of these queries that fall inside the evaluation image and pass the rendered-alpha threshold. 
It measures evaluation support rather than reconstruction completeness or accuracy. 
\emph{Robot Error} compares the 3DGS-rendered depth at
kinematically projected robot-surface queries with the camera-axis depth predicted by forward kinematics and CAD. 
\emph{Stereo Disagreement} compares rendered depth with confidence-filtered SGBM depth over their jointly valid pixels.

Pose conditioning alone does not consistently improve these measurements.
With shipped DROID extrinsics, pose conditioning slightly lowers both
YoNoSplat errors. For ZipSplat, the aggregate robot-error median rises from
$19.71$ to $30.58$\,cm while stereo disagreement decreases. However, the
paired robot-error change is not significant (median $\Delta=-1.10$\,cm,
$p=0.072$, $n=109$). Thus, the aggregate increase does not establish
systematic degradation but reflects heterogeneous scene-level responses.
With refined extrinsics, aggregate errors decrease for ZipSplat and remain
nearly unchanged for YoNoSplat. Robot projection coverage is identical
between pose-free and pose-conditioned reconstruction within each extrinsic
source, so these differences are measured over the same projected support.

\subsubsection{Qualitative Results}

\begin{table*}[t]
  \centering
  \small
  \setlength{\tabcolsep}{5pt}
  \renewcommand{\arraystretch}{0.95}
  \newcommand{\yes}{\ding{51}}
  \newcommand{\no}{\ding{55}}
  \begin{tabular}{ll cc rrr}
    \toprule
    \multirow{2}{*}{Backbone}
    & \multirow{2}{*}{Variant}
    & \multirow{2}{*}{Pose Inj.}
    & \multirow{2}{*}{Refined Ext.}
    & \multicolumn{3}{c}{Photometric Fidelity (3-view)} \\
    \cmidrule(lr){5-7}
    & & & &
    PSNR$\uparrow$ & SSIM$\uparrow$ & LPIPS$\downarrow$ \\
    \midrule

    \multirow{2}{*}{AnySplat~\cite{jiang2025anysplat}}
    & Baseline
      & \no & \no
      & 8.67 & 0.3707 & 0.6641 \\
    & Baseline + refined ext.
      & \no & \yes
      & \textbf{9.60} & \textbf{0.4037} & \textbf{0.5911} \\

    \midrule
    \multirow{4}{*}{YoNoSplat~\cite{yonosplat}}
    & Baseline
      & \no & \no
      & 12.57 & 0.3582 & 0.5588 \\
    & Baseline + pose inj.
      & \yes & \no
      & 15.26 & 0.4831 & 0.3074 \\
    & Baseline + refined ext.
      & \no & \yes
      & 14.45 & 0.4240 & 0.3187 \\
    & Ours
      & \yes & \yes
      & \textbf{15.81} & \textbf{0.5105} & \textbf{0.2592} \\

    \midrule
    \multirow{4}{*}{ZipSplat~\cite{veicht2026zipsplat}}
    & Baseline
      & \no & \no
      & 12.75 & 0.4233 & 0.5467 \\
    & Baseline + pose inj.
      & \yes & \no
      & 14.50 & 0.4775 & 0.4437 \\
    & Baseline + refined ext.
      & \no & \yes
      & 15.19 & 0.4972 & 0.3464 \\
    & Ours
      & \yes & \yes
      & \textbf{16.27} & \textbf{0.5242} & \textbf{0.2979} \\

    \bottomrule
  \end{tabular}

  \caption{
  \textbf{Ablation of pose conditioning and refined extrinsics on up to 110 scenes.} 
  Baseline uses shipped DROID extrinsics without pose conditioning.
  \emph{Baseline + pose injection} conditions the backbone on the shipped
  poses; \emph{Baseline + refined extrinsics} uses PointWorld-refined
  extrinsics while keeping the backbone pose-free; Ours combines refined
  extrinsics with pose conditioning. AnySplat ($n=107$) is reported only for compatible
  pose-free configurations because it does not accept external camera poses.
  }
  \label{tab:photometric_ablation}
\end{table*}

\begin{figure*}[t]
  \centering

 \includegraphics[width=1.0\linewidth]{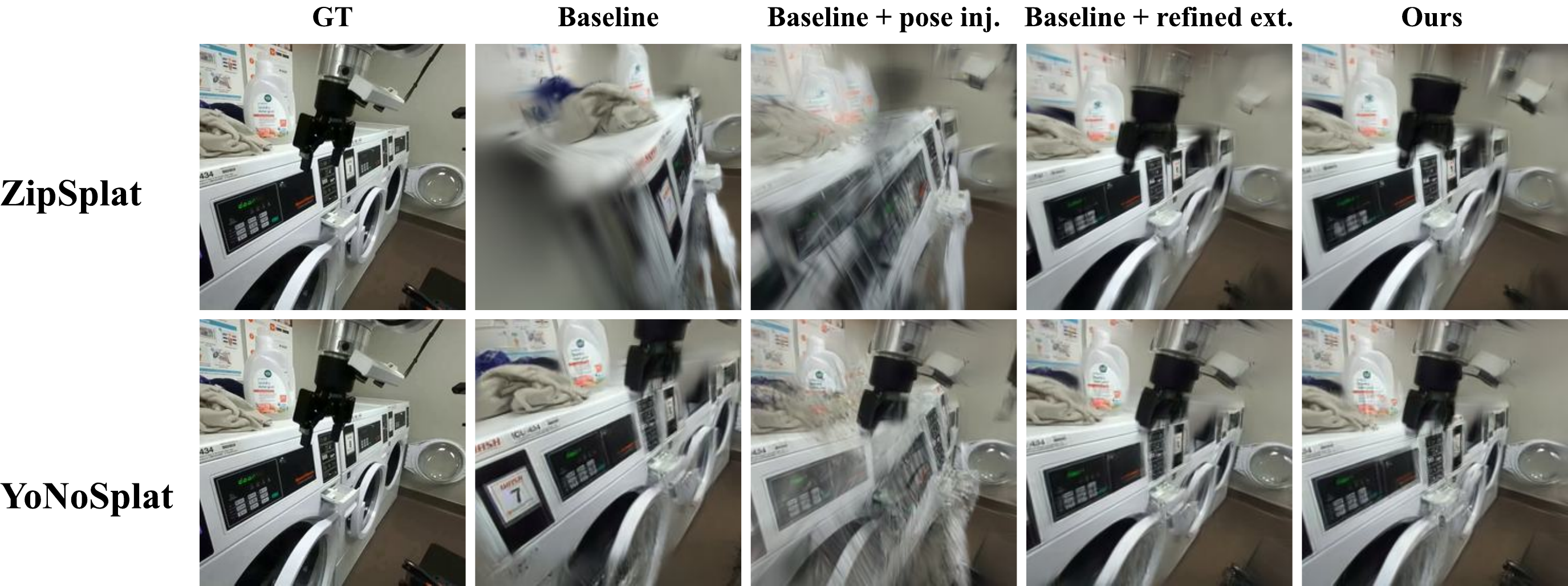}
  \caption{\textbf{Qualitative ablation on held-out \texttt{ext2R}.}
Rows show ZipSplat and YoNoSplat on the same scene. After the ground-truth
column, reconstruction columns follow the variants defined in
Table~\ref{tab:photometric_ablation}. All variants use the same three exterior
input views.}
  \label{fig:qualitative_ablation}
\end{figure*}

\paragraph{Qualitative Results.}
Figure~\ref{fig:qual} compares the pose-free DROID baseline with Ours, which combines refined extrinsics and pose conditioning. 
All scenes are reconstructed from three exterior views and rendered at the held-out \texttt{ext2R} camera. 
Across the examples, the baseline often exhibits viewpoint misalignment and scene distortion, whereas Ours more closely matches the ground-truth viewing direction and produces fewer visible artifacts.
Because both the extrinsic source and pose conditioning change in this comparison, the figure demonstrates their combined effect rather than attributing the improvement to either component.

\subsubsection{Ablation Studies}

\paragraph{Quantitative Ablation.}
Table~\ref{tab:photometric_ablation} separates the effects of pose
conditioning and refined extrinsics using paired per-scene differences. At a
fixed extrinsic source, pose conditioning improves all three photometric
metrics for both supported backbones, with \emph{median per-scene PSNR} gains ranging
from $+0.73$ to $+2.73$\,dB (all PSNR $p<10^{-10}$). At a fixed pose setting,
switching to refined extrinsics yields median PSNR gains of $+0.29$ to
$+2.13$\,dB and likewise improves SSIM and LPIPS (all $p<2\times10^{-8}$). Combining both components performs
best for YoNoSplat and ZipSplat, although their gains are non-additive. 

Of the 110 valid-intrinsic scenes, AnySplat is evaluated on 107 because its
pose-free camera estimator produces degenerate camera layouts in three cases,
preventing reliable alignment to the metric camera rig. AnySplat gains $+0.39$\,dB under the same shipped-to-refined contrast. This change reflects scene registration and held-out camera
placement rather than pose conditioning. 

\paragraph{Qualitative Ablation.}
As shown in Figure~\ref{fig:qualitative_ablation}, the two DROID-calibrated results
exhibit scene distortion and viewpoint misalignment relative to the ground
truth, whereas the refined-extrinsic results recover a more consistent viewing
angle. For ZipSplat, the four settings achieve PSNR values of $9.81$, $10.85$,
$14.86$, and $14.97$\,dB, respectively. Despite the similar scores of the last
two settings, the pose-free result exposes more of the washing machine's left
panel, while Ours more closely matches the ground-truth framing. YoNoSplat
shows the same overall tendency, with PSNR increasing from $8.49$\,dB at
Baseline to $15.73$\,dB with Ours. This comparison supports the quantitative
ablation while revealing local viewpoint errors that PSNR alone may obscure.

\begin{wraptable}{r}{0.49\textwidth}
  \centering
  \small
  \setlength{\tabcolsep}{3pt}
  \renewcommand{\arraystretch}{0.95}
  \begin{tabular}{c rr c rr}
    \toprule
    \multicolumn{3}{c}{Ratio-band sweep}
    & \multicolumn{3}{c}{Support sweep} \\
    \cmidrule(lr){1-3}\cmidrule(lr){4-6}
    Band & Ref. & Ship.
    & $n_{\min}$ & Ref. & Ship. \\
    \midrule
    $\pm 5\%$  & 55  & 40 & 1   & 138 & 88 \\
    $\pm 10\%$ & 99  & 69 & 10  & 127 & 82 \\
    $\pm 15\%$ & \textbf{114} & \textbf{80}
               & 30  & \textbf{114} & \textbf{80} \\
    $\pm 20\%$ & 122 & 89 & 50  & 103 & 68 \\
    $\pm 30\%$ & 122 & 91 & 100 & 91  & 60 \\
    $\pm 50\%$ & 125 & 94 & 300 & 53  & 38 \\
    \bottomrule
  \end{tabular}
  \caption{\textbf{Gate-threshold sensitivity over the 225 eligible scenes.}
  The ratio-band sweep fixes $n_{\min}=30$, while the support sweep fixes
  the band to $\pm15\%$. Bold denotes the release setting.}
  \label{tab:gate_sensitivity}
\end{wraptable}

\paragraph{Gate-Threshold Sensitivity.}
Table~\ref{tab:gate_sensitivity} shows that 114 is determined by the selected
operating point rather than being an intrinsic dataset size. The pass count
nearly saturates beyond a $\pm20\%$ ratio band, whereas changing the minimum
support has a larger effect, indicating that scene selection is governed
primarily by robot observability. Refined extrinsics yield more passing scenes
than shipped extrinsics under every tested setting, although this ordering
measures gate consistency rather than independent absolute pose accuracy.

\subsubsection{Robustness Checks}
\label{sec:notartifact}

The calibration gain extends beyond the probe comparing robot geometry with depth. 
Although PointWorld does not optimize rendering fidelity~\cite{huang2026pointworld},
refined extrinsics yield median paired PSNR gains of $+2.13$, $+1.42$, and
$+0.39$\,dB for ZipSplat, YoNoSplat, and AnySplat ($n=110,110,107$), with
SSIM and LPIPS agreeing. Because AnySplat does not consume poses, its fixed Gaussians isolate
improvements in registration between the camera and the scene.
A secondary \texttt{ext1L} test split by the median maximum left-camera correction
($11.23$\,cm; $n=109$) yields $+0.89$\,dB for the larger-correction half and
$+0.00$\,dB for the smaller half.

Photometry still does not establish metric placement. Under shipped
extrinsics without pose conditioning, robot-depth ratios are $1.34$, $1.31$,
and $1.26$ for the same backbones, consistently placing the arm too far.
This motivates our robot-referenced metric and per-scene reliability
annotations.

\section{Discussion and Limitations}

3DROID currently reconstructs a single static frame from four exterior views
and measures geometry only where robot-surface or stereo support is available.
Additional well-calibrated views may reduce sparse-view ambiguity, but we do
not isolate this effect. We exclude the wrist stream because its narrow field
of view and large domain gap from stationary cameras are
poorly matched to current feed-forward backbones. 
Adapting these models to mixed observations from exterior and wrist cameras,
or fine-tuning them on robot data with metric pose, stereo, kinematic, and CAD
supervision, is promising future work.
A natural extension is to expand 3DROID from static snapshots to temporally
consistent 4D scenes that could provide dynamic context to robot learning
models.
We do not claim full-scene geometric ground truth or downstream manipulation
improvements.

\section{Conclusion}
To complement 2D-centric robot manipulation data with explicit 3D scene information, 
we studied how to construct 3D Gaussian representations anchored to the metric scale and coordinate frame of real-world workspaces. 
Our experiments show that pose conditioning constrains global similarity ambiguity, 
while its outcome depends critically on the reliability of the injected extrinsics and does not alone guarantee metric geometry. 
We therefore assess physical consistency separately through coverage-aware agreement between rendered depth, 
classical stereo, and kinematically localized robot geometry. 
Based on these findings, we introduce 3DROID, a metric-pose-anchored dataset that provides renderable robot scenes together with measured calibration and reconstruction-reliability information.

\bibliographystyle{plain}  %
\bibliography{sec/references} %

\clearpage
\appendix

\section{Additional experimental details} %

\subsection{Compute resources and runtime} %
\label{sec:compute_resource}

All experiments used a single NVIDIA A100 80GB GPU. Model loading took
20.7\,s, and the first scene required approximately 250\,s for one-time CUDA
kernel compilation. Excluding these initialization costs, reconstruction took
a median of 0.77\,s per scene across all 114 released scenes. The reported
compute also includes preliminary and repeated experiments conducted during
model selection, ablation, and debugging.

\subsection{Released Code and Reproducibility}

The released dataset provides the 114 episode identifiers, a pinned software
environment, the unmodified ZipSplat source, checkpoint revision and checksum,
and a resumable pipeline for rebuilding the released four-view Gaussians and
depth maps from externally obtained DROID videos and camera metadata. It also
reconstructs the three-view held-out \texttt{ext2R} renderings under the four
pose--extrinsic conditions. Running
\texttt{python pipeline/run\_pipeline.py all} performs frame extraction,
reconstruction, rendering, and verification. On our reference A100 environment,
all 114 scenes reproduce the stored depth, alpha, camera, and metric-scale
arrays exactly. PLY hashes are reported separately because atomic rasterization
need not be byte-identical across hardware.

For verification without raw videos or a GPU,
\texttt{python evaluation/make\_tables.py --bundle .} re-aggregates the released
per-scene measurements and reproduces the ZipSplat entries in
Tables~\ref{tab:photometric}, \ref{tab:geometric}, and
\ref{tab:photometric_ablation}, including their sample counts. This operation
verifies the reported aggregation rather than re-measuring geometry. The
current release does not rerun the YoNoSplat and AnySplat baselines. The required external inputs are documented in the release README.

\section{Asset Licenses and Data Availability}
\label{sec:asset_licenses}

\paragraph{Existing assets.}
We use publicly released datasets and pretrained reconstruction models under
their respective terms. DROID~\cite{khazatsky2024droid}, including the source
images, robot states, intrinsics, and shipped extrinsics, is released under
CC BY 4.0. We redistribute only the camera parameters required to use 3DROID;
the source images and robot states remain in DROID and are joined through the
episode UUID.
We use the \texttt{droid\_raw} v1.0.1 release, accessed at
\url{gs://gresearch/robotics/droid_raw/1.0.1/}.

The refined left-eye extrinsics are obtained from
PointWorld-DROID~\cite{huang2026pointworld} under the NVIDIA License. We use the PointWorld-DROID release at revision
\texttt{dd9aaeec94bb14e27ab6b16b6e4aa0dbcf3ef56f}. The
corresponding right-eye poses are derived using the original within-rig stereo
transform. In accordance with Sections 3.1--3.3 of that license, we include the
complete license, preserve upstream attribution, identify the derived assets,
and retain its restriction to non-commercial research or evaluation.

\paragraph{Reconstruction models.}
The released reconstructions use the official ZipSplat checkpoint
\texttt{zipsplat-da3g-252p.tar}~\cite{veicht2026zipsplat}. ZipSplat code is
licensed under Apache-2.0 and its pretrained weights under CC BY-NC 4.0.
YoNoSplat~\cite{yonosplat} and AnySplat~\cite{jiang2025anysplat} are used only
as experimental baselines; their official code is licensed under MIT. Model
code and pretrained weights are obtained from the original repositories and
are not redistributed in 3DROID.

\paragraph{3DROID release.}
The 114-scene 3DROID dataset is publicly available during review at
\url{https://huggingface.co/datasets/wonguen/3DROID}. Documentation
and author-provided release material are licensed under Apache-2.0. To the
extent that we hold the relevant rights, the generated Gaussian, depth, and
measurement assets are released under CC BY-NC 4.0, while remaining subject
to applicable upstream terms. Camera metadata retains its component-specific
DROID and PointWorld terms. The repository includes the corresponding license
files and third-party notices and will be de-anonymized in the camera-ready
release.

The repository additionally includes the reconstruction and verification
scripts, pinned dependencies, episode list, per-file manifest, and numerical
reproduction records.

The release vendors only the unmodified ZipSplat source under Apache-2.0.
YoNoSplat and AnySplat code and checkpoints are not included; their official
repositories, revisions, and checkpoint hashes are provided for provenance.
\clearpage

\end{document}